\documentclass[10pt]{article}

\usepackage[margin=1in]{geometry}
\usepackage{times}
\usepackage{microtype}
\usepackage{amsmath,amssymb}
\usepackage{booktabs}
\usepackage{enumitem}
\usepackage{hyperref}
\usepackage{url}
\usepackage{algorithm}
\usepackage{algpseudocode}

\usepackage{algorithm}
\usepackage{algpseudocode}
\usepackage{microtype}
\usepackage{float} 
\usepackage{graphicx}

\title{\vspace{-0.8em}Significance-Gain Pair Encoding for LLMs:\\ A Statistical Alternative to Frequency-Based Subword Merging\vspace{-0.5em}}
\author{Azam Nouri\\\small (Department of Science, Technology \& Mathematics, Lincoln University
)}
\date{2026}

\begin{document}
\maketitle

\begin{abstract} Subword tokenization is a key design choice for modern language models, including large language models (LLMs), with byte/character-level BPE serving as a widely used baseline. Standard BPE selects merges by raw pair frequency, which favors compression but can conflate true adjacency cohesion with pairs that are frequent due to high marginal counts. We introduce \emph{Significance-Gain BPE}, a drop-in alternative merge criterion that measures cohesion via a z-statistic under an independence null model and combines it with an explicit compression-aware gain term. We evaluate Significance-Gain BPE on WikiText-103 (raw) character slices using a small causal Transformer language model and report both token-dependent perplexity and the tokenizer-invariant metric bits-per-character (BPC). At a representative operating point, Significance-Gain BPE reduces validation and test perplexity by 13\% and 12\%, respectively, and improves validation and test BPC by $\sim$0.9--1.0\%. A vocabulary-size sweep further shows lower BPC in most closest-compression comparisons, suggesting that statistically grounded merge selection can improve predictive efficiency per unit of raw text across a range of compression regimes. \end{abstract}

\section{Introduction}
Subword tokenizers (BPE, WordPiece, Unigram/SentencePiece) are the standard interface between raw text and neural language models \cite{sennrich2016bpe,schuster2012wordpiece,kudo2018sentencepiece}. Tokenization affects (i) sequence length (hence context utilization and training efficiency), (ii) vocabulary size and softmax cost, and (iii) the statistical structure presented to the model. While byte-level language models can remove tokenization entirely \cite{xue2021byt5}, byte/character-level BPE remains attractive due to its simplicity, language agnosticism, and robust handling of rare strings \cite{radford2019gpt2}.

\paragraph{Motivation.}
Standard BPE chooses merges by raw frequency: it repeatedly replaces the most common adjacent pair. This greedy rule strongly favors \emph{compression}: merging a pair that appears $c_{xy}$ times reduces the token sequence length by approximately $c_{xy}$ at that iteration. However, frequency alone can be misleading because it does not distinguish:
(i) pairs that are frequent merely because their constituents are individually common (high marginals), from
(ii) pairs that are frequent because they form a cohesive unit (strong adjacency association). 
In corpora with heavy whitespace, punctuation, and common affixes, frequency-driven merges can over-prioritize ``background'' adjacency patterns and under-prioritize informative collocations.

\paragraph{Core idea: cohesion $\times$ gain.}
The proposed method separates cohesion from compression. Let $c_x$ and $c_y$ denote symbol counts and let $c_{xy}$ denote adjacent pair counts over $N$ adjacent positions. Under an independence null model, the expected adjacency count is
\begin{equation}
\mathbb{E}[c_{xy}] \;=\; \frac{c_x c_y}{N}.
\label{eq:expected_intro}
\end{equation}
More-than-expected adjacency is measured using the z-statistic
\begin{equation}
z(x,y) \;=\; \frac{c_{xy} - \mathbb{E}[c_{xy}]}{\sqrt{\mathbb{E}[c_{xy}] + \varepsilon}},
\label{eq:z_intro}
\end{equation}
and combine it with an explicit compression gain term (proportional to $c_{xy}$) to score merges. This yields a greedy algorithm that prefers pairs that are both (i) statistically cohesive and (ii) compressive.

\paragraph{Evaluation: token-dependent vs.\ token-invariant metrics.}
Per-token perplexity (PPL) and NLL depend on the tokenizer because ``a token'' changes as merges change. Accordingly, both PPL and a tokenizer-invariant metric derived from NLL per character are reported:
\begin{equation}
\mathrm{NLL/char} = \mathrm{NLL/token}\cdot \mathrm{TPC}, 
\qquad
\mathrm{BPC} = \frac{\mathrm{NLL/char}}{\ln 2},
\end{equation}
where $\mathrm{TPC}$ is tokens-per-character. BPC allows fair comparison across tokenizers by normalizing log-likelihood by raw text length.

\paragraph{Contributions.}
\begin{itemize}[leftmargin=*,itemsep=0.3em]
\item \textbf{Significance-Gain BPE.} A merge objective that combines standardized adjacency surprise under an independence null with an explicit compression gain term (Sec.~2).
\item \textbf{Minimal implementation and diagnostics.} A lightweight tokenizer+LM pipeline that computes both token-dependent (PPL) and token-invariant (BPC) metrics, and supports matched-compression sweeps (Sec.~3--4).
\item \textbf{Empirical results.} On WikiText-103 character slices, Significance-Gain BPE improves BPC by about 0.9--1.0\% at a representative operating point (and improves PPL by $\approx$12--13\%), with grid results indicating mostly consistent gains at matched compression (Sec.~4).
\end{itemize}

\paragraph{Paper organization.}
Sec.~2 defines the scoring rule and contrasts it with frequency BPE. Sec.~3 describes the experimental pipeline. Sec.~4 reports results, including matched-compression comparisons and ablations. 


\section{Significance-Gain BPE vs.\ Standard Frequency BPE}
\label{sec:sig_gain_bpe}

\subsection{Standard BPE as a compression-driven greedy algorithm}
Byte Pair Encoding (BPE) was originally introduced as a data-compression procedure: repeatedly replace the most common adjacent symbol pair with a new symbol \cite{gage1994bpe}. Modern subword BPE for NLP follows the same greedy merge principle \cite{sennrich2016bpe}. Given a token sequence $s=(s_1,\dots,s_T)$, define the adjacent pair count
\begin{equation}
c_{xy} \;=\; \sum_{t=1}^{T-1}\mathbf{1}[s_t=x,\, s_{t+1}=y],
\end{equation}
and choose $(x^\star,y^\star)=\arg\max_{(x,y)} c_{xy}$ (often subject to $c_{xy}\ge c_{\min}$). Replacing all occurrences of $(x^\star,y^\star)$ by a merged symbol reduces the token sequence length by approximately $c_{x^\star y^\star}$, so frequency-BPE is a strong heuristic for maximizing \emph{immediate compression}.

\paragraph{Limitation (marginals vs.\ cohesion).}
Raw frequency conflates two effects:
(i) $(x,y)$ can be frequent because $x$ and $y$ are individually frequent (high marginals), and
(ii) $(x,y)$ can be frequent because $x$ and $y$ form a cohesive unit (strong association).
In practice, this can bias merges toward background adjacency patterns (e.g., whitespace and punctuation) rather than statistically meaningful collocations.

\subsection{Independence null model and statistical cohesion}
Let $c_x=\sum_{t}\mathbf{1}[s_t=x]$ be symbol counts and let $N=\max(T-1,1)$ be the number of adjacency positions. Under an independence null model, the expected adjacency count is
\begin{equation}
\mathbb{E}[c_{xy}] \;=\; \frac{c_x c_y}{N}.
\label{eq:expected}
\end{equation}
The extent of more-than-expected adjacency is quantified using a standardized residual (z-score):
\begin{equation}
z(x,y) \;=\; \frac{c_{xy} - \mathbb{E}[c_{xy}]}{\sqrt{\mathbb{E}[c_{xy}] + \varepsilon}}.
\label{eq:zscore}
\end{equation}
This form is closely related to Pearson residuals in contingency tables and is consistent with classical collocation statistics that separate association from marginal frequency (e.g., mutual information and likelihood-ratio tests) \cite{church1990mi,dunning1993llr}.

\paragraph{Connection to PMI (intuition).}
Pointwise mutual information can be written (in counts) as
\begin{equation}
\mathrm{PMI}(x,y) = \log\frac{c_{xy}N}{c_x c_y}.
\end{equation}
Using $c_{xy} = \mathbb{E}[c_{xy}]\,e^{\mathrm{PMI}}$, the numerator in \eqref{eq:zscore} becomes
$c_{xy}-\mathbb{E}[c_{xy}] = \mathbb{E}[c_{xy}]\,(e^{\mathrm{PMI}}-1)$,
so
\begin{equation}
z(x,y) \;\approx\; \sqrt{\mathbb{E}[c_{xy}]}\,\big(e^{\mathrm{PMI}(x,y)}-1\big).
\end{equation}
Thus, unlike PMI alone, the z-score naturally incorporates support through $\sqrt{\mathbb{E}[c_{xy}]}$, reducing the tendency of extremely rare pairs to dominate.

\subsection{Significance-Gain score used in the implementation}
\label{sec:sig_gain_score}

\paragraph{Counts and notation.}
Given a token sequence $s=(s_1,\dots,s_T)$ (characters at initialization, then merged symbols), define:
\begin{itemize}[leftmargin=*,itemsep=0.2em]
\item $c_x = \sum_{t=1}^{T}\mathbf{1}[s_t=x]$: count of symbol $x$ in the current sequence,
\item $c_{xy} = \sum_{t=1}^{T-1}\mathbf{1}[s_t=x,\,s_{t+1}=y]$: count of adjacent pair $(x,y)$,
\item $N = \max(T-1,1)$: number of adjacent positions in the sequence.
\end{itemize}
Only candidate pairs with $c_{xy}\ge c_{\min}$ are considered.
\paragraph{Cohesion (significance) term.}
Under an independence null model (successor determined by marginals), the expected adjacency count of $(x,y)$ is
\begin{equation}
\mathbb{E}[c_{xy}] \;=\; \frac{c_x c_y}{N}.
\end{equation}
The extent to which the observed adjacency is ``more-than-expected'' is measured using the z-statistic
\begin{equation}
z(x,y) \;=\; \frac{c_{xy} - \mathbb{E}[c_{xy}]}{\sqrt{\mathbb{E}[c_{xy}] + \varepsilon}},
\end{equation}
where $\varepsilon>0$ is a small constant for numerical stability. Large positive $z(x,y)$ indicates that $x$ is followed by $y$ more often than explained by their marginal counts, i.e., the pair is cohesive.

\paragraph{Compression (gain) term.}
A merge replaces each occurrence of adjacent $(x,y)$ by a single symbol, reducing sequence length by approximately one token per occurrence. Therefore the immediate compression benefit is proportional to $c_{xy}$.vThis quantity is referred to as the \emph{gain}:
\begin{equation}
\mathrm{gain}(x,y) \;\approx\; c_{xy}.
\end{equation}

\paragraph{Final merge score (``sigz'' mode) and the \texttt{use\_gain} switch.}
The implementation supports two scoring variants: a \emph{gain-weighted} score (default) and a \emph{cohesion-only} score used for ablation. This is controlled by the Boolean flag \texttt{use\_gain}.

\begin{itemize}[leftmargin=*,itemsep=0.2em]
\item If \texttt{use\_gain=True}, the score is multiplied by $c_{xy}$ so that pairs that occur many times (and thus yield larger immediate compression) are preferred.
\item If \texttt{use\_gain=False}, the score omits this multiplicative $c_{xy}$ factor, isolating the effect of statistical cohesion (useful for ablations and diagnostics).
\end{itemize}

The implementation combines cohesion and gain as:
\begin{equation}
\mathrm{score}(x,y)
\;=\;
\underbrace{\Big(\mathbf{1}[\texttt{use\_gain}]\,c_{xy}+\mathbf{1}[\neg\texttt{use\_gain}]\Big)}_{\text{gain switch}}
\cdot
\underbrace{\Big(z(x,y)\cdot c_{xy}^{\alpha}\Big)}_{\text{support-tempered cohesion}}
\;-\;
\underbrace{\lambda_{\mathrm{rare}}(c_{xy}+\varepsilon)^{-1/2}}_{\text{optional rare-merge penalty}}.
\label{eq:final_score}
\end{equation}
Here:
\begin{itemize}[leftmargin=*,itemsep=0.2em]
\item $\mathbf{1}[\cdot]$ is an indicator (1 if the condition is true, else 0).
\item $\alpha\in[0,1]$ (\texttt{alpha\_count}) controls how strongly additional support boosts the cohesion term:
$\alpha=0$ uses only $z(x,y)$, while $\alpha>0$ increasingly prefers higher-count pairs among those with high $z$.
\item $\lambda_{\mathrm{rare}}\ge 0$ is a penalty weight that discourages selecting very rare pairs even if their $z$ is large.
\end{itemize}
In the main configuration used for the reported results, \texttt{use\_gain=True}, $\alpha=0.25$, $\lambda_{\mathrm{rare}}=0$, and $c_{\min}=5$.

\paragraph{Interpretation.}
The score favors pairs that are (i) statistically cohesive (large positive $z$), and (ii) compressive (large $c_{xy}$). The $c_{xy}^{\alpha}$ factor stabilizes the selection by favoring merges with reliable count support, while the optional rare penalty provides an additional safeguard against overly sparse merges. Overall, the objective is a greedy compromise between association-driven merging and compression-driven merging.

\paragraph{Interpretation.}
The score favors pairs that (i) are surprisingly frequent relative to marginals (large $z$), (ii) occur enough times to matter for compression (via $c_{xy}$), and (iii) are not too sparse (via $c_{\min}$ and optional $\lambda_{\mathrm{rare}}$). This objective can be viewed as a greedy compromise between association-driven merging and compression-driven merging.

\subsection{Training and encoding}
The training and encoding procedures are identical to standard BPE except for the pair-selection rule:

\begin{algorithm}[t]
\caption{Significance-Gain BPE training (character-base)}
\label{alg:sig_gain_bpe}
\begin{algorithmic}[1]
\Require Training text $\mathcal{D}$; minimum pair count $c_{\min}$; target vocabulary size $V_{\text{target}}$; stability constant $\varepsilon$
\State $s \gets$ list of characters in $\mathcal{D}$
\State $M \gets [\ ]$ \Comment{ordered list of merges}
\While{$|\mathcal{V}(s)| < V_{\text{target}}$}
  \State Compute symbol counts $\{c_x\}$ on $s$
  \State Compute adjacent pair counts $\{c_{xy}\}$ on $s$
  \State $\mathcal{C} \gets \{(x,y) : c_{xy} \ge c_{\min}\}$
  \If{$\mathcal{C}=\emptyset$}
    \State \textbf{break} \Comment{no admissible merges remain}
  \EndIf
  \State $(a,b) \gets \arg\max\limits_{(x,y)\in\mathcal{C}} \mathrm{score}(x,y)$
  \State \hspace{1.2em}\Comment{$\mathrm{score}$ defined in Eq.~\eqref{eq:final_score}}
  \State Create merged symbol $m \gets ab$
  \State Replace all occurrences of adjacent $(a,b)$ in $s$ with $m$
  \State Append $((a,b),m)$ to $M$
\EndWhile
\State \Return merge list $M$ and vocabulary $\mathcal{V}(s)$
\end{algorithmic}
\end{algorithm}


\section{Experimental Setup and Evaluation Pipeline}
\label{sec:exp_setup}

This section describes the end-to-end pipeline used in the provided notebook: (i) train a tokenizer, (ii) encode train/val/test, (iii) train a small causal LM, and (iv) compare language-model metrics using both token-dependent and token-invariant measures.

\subsection{Data}
WikiText-103 (raw) is used with lightweight whitespace normalization to stabilize character-level encoding. For rapid iteration, contiguous character slices are taken:
\begin{equation}
|\mathcal{D}_{\text{train}}|=1{,}000{,}000,\quad
|\mathcal{D}_{\text{val}}|=200{,}000,\quad
|\mathcal{D}_{\text{test}}|=200{,}000
\;\;\text{characters.}
\end{equation}

\subsection{Base tokenization and BPE training}
To ensure nonzero merges and avoid confounds from large pretokenized vocabularies, the reported experiments use a \textbf{character-base} initialization: the initial sequence is the list of characters in the training text. Both tokenizers then apply the same training loop, differing only in the pair-selection rule (frequency vs.\ Eq.~\eqref{eq:final_score}).

\paragraph{Tokenizer training outcome (main configuration).}
With the main configuration (Significance-Gain with \texttt{use\_gain=True}, $\alpha=0.25$, $\lambda_{\mathrm{rare}}=0$), the training logs show:
\begin{itemize}[leftmargin=*,itemsep=0.2em]
\item Frequency-BPE: 469 merges, final vocabulary size 617.
\item Significance-Gain BPE: 487 merges, final vocabulary size 635.
\end{itemize}
This indicates both tokenizers perform substantial merging beyond the character base.

\subsection{Tokenizer-level metrics}
The following metrics are reported:
\begin{itemize}[leftmargin=*,itemsep=0.2em]
\item \textbf{Tokens-per-character (TPC)}:
\begin{equation}
\mathrm{TPC}(\mathcal{D}) \;=\; \frac{\#\text{tokens}(\mathrm{encode}(\mathcal{D}))}{\#\text{characters}(\mathcal{D})},
\end{equation}
which quantifies compression (lower is shorter sequences).
\item \textbf{Vocabulary utilization}:
\begin{equation}
\mathrm{vocab\_used}(\mathcal{D}) \;=\; \frac{|\mathrm{unique}(\mathrm{encode}(\mathcal{D}))|}{|\mathcal{V}|},
\end{equation}
which measures how much of the trained vocabulary is active on each split.
\end{itemize}

\paragraph{Observed tokenizer statistics (main configuration).}
For frequency-BPE, TPC is 0.4364 (val) and 0.4307 (test). For Significance-Gain BPE, TPC is 0.4430 (val) and 0.4366 (test). Thus, Significance-Gain BPE yields slightly longer token sequences, motivating the use of token-invariant evaluation (Sec.~\ref{sec:lm_metrics}).

\subsection{Language model and training}
A small causal Transformer (\texttt{TinyGPT}) is trained with embedding dimension $d=192$, $L=4$ Transformer blocks, and $h=4$ attention heads. Training uses block size 256 with stride 256, AdamW with learning rate $3\times 10^{-4}$, batch size 64, and 8 epochs. Validation is evaluated each epoch and final metrics are reported on both val and test.

\subsection{LM metrics}
\label{sec:lm_metrics}
Token-level negative log-likelihood (NLL), perplexity, and token-level accuracy are reported:
\begin{equation}
\mathrm{PPL}=\exp(\mathrm{NLL}).
\end{equation}
Because per-token metrics depend on how the text is segmented into tokens, a tokenizer-invariant metric derived from NLL per character is additionally reported:
\begin{equation}
\mathrm{NLL/char} = \mathrm{NLL/token}\cdot \mathrm{TPC},
\qquad
\mathrm{BPC} = \frac{\mathrm{NLL/char}}{\ln 2}.
\end{equation}
BPC (bits-per-character) enables fair comparison across tokenizers by normalizing log-likelihood by the length of the original text rather than the number of tokens.

\subsection{Matched-compression sweep}
Beyond the main operating point, a grid of vocabulary sizes $V\in\{300,400,600,800,1200\}$ is evaluated and validation BPC is plotted versus validation TPC. For each Significance-Gain configuration, a ``closest-TPC'' frequency-BPE comparison is also reported to control for compression differences. These results are summarized in Fig.~\ref{fig:bpc_tpc} and Table~\ref{tab:matched_compression} in the Results section (Sec.~4).


\section{Results}
\label{sec:results}

\subsection{Main operating point}
Frequency-BPE and Significance-Gain BPE are first compared under a matched LM training budget (Sec.~\ref{sec:exp_setup}). Table~\ref{tab:main_results} reports per-token perplexity and the tokenizer-invariant bits-per-character (BPC). While Significance-Gain BPE achieves a large reduction in token-level perplexity, the more appropriate cross-tokenizer comparison is BPC, which normalizes by raw character length.

\begin{table}[t]
\centering
\begin{tabular}{@{}lcccc@{}}
\toprule
Tokenizer & Val PPL $\downarrow$ & Test PPL $\downarrow$ & Val BPC $\downarrow$ & Test BPC $\downarrow$ \\
\midrule
Standard (freq) BPE & 312.78 & 323.71 & 3.6176 & 3.5915 \\
Significance-Gain BPE ($\alpha{=}0.25$) & 271.47 & 283.85 & 3.5818 & 3.5580 \\
\midrule
Improvement (\%) & 13.21 & 12.31 & 0.99 & 0.93 \\
\bottomrule
\end{tabular}
\caption{Main operating point (character-base). Per-token perplexity decreases substantially, while the tokenizer-invariant BPC improves by about 0.9--1.0\%. Significance-Gain BPE yields slightly higher TPC (0.4430 vs.\ 0.4364 on validation), so BPC is the appropriate primary metric.}
\label{tab:main_results}
\end{table}

\paragraph{Interpretation.}
Significance-Gain BPE improves NLL per token, leading to large PPL reductions. However, it also yields slightly higher tokens-per-character (TPC), i.e., longer token sequences. BPC integrates both effects and therefore provides a fair comparison across tokenizers. The observed BPC improvements (0.99\% on validation and 0.93\% on test) indicate that the proposed merge objective improves predictive efficiency per unit of raw text.

\subsection{Compression--quality trade-off: BPC vs.\ TPC}
\begin{figure}[t]
\centering
\includegraphics[width=0.85\linewidth]{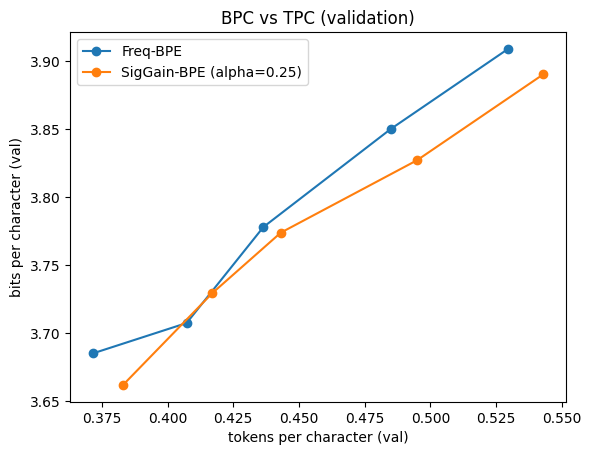}
\caption{Validation bits-per-character (BPC) versus tokens-per-character (TPC) across a vocabulary-size sweep. Lower is better on both axes. Significance-Gain BPE (alpha=0.25) is typically below frequency-BPE, indicating improved predictive efficiency per unit of text across most compression regimes.}
\label{fig:bpc_tpc}
\end{figure}

To understand how the effect varies across operating points, vocabulary sizes $V\in\{300,400,600,800,1200\}$ are swept and validation BPC is plotted versus validation TPC (Fig.~\ref{fig:bpc_tpc}). For each Significance-Gain point, a comparison is also made against the closest-TPC frequency-BPE point (Table~\ref{tab:matched_compression}).

\begin{table}[t]
\centering
\begin{tabular}{@{}rcccc@{}}
\toprule
V & TPC (SigGain) & BPC (SigGain) & TPC (Freq) & $\Delta$BPC (Freq -- SigGain) \\
\midrule
300  & 0.5430 & 3.8903 & 0.5295 & +0.0186 \\
400  & 0.4949 & 3.8271 & 0.4848 & +0.0228 \\
600  & 0.4430 & 3.7739 & 0.4364 & +0.0042 \\
800  & 0.4170 & 3.7295 & 0.4072 & -0.0220 \\
1200 & 0.3831 & 3.6621 & 0.3715 & +0.0231 \\
\bottomrule
\end{tabular}
\caption{Matched-compression comparison on validation: each Significance-Gain point is paired with the closest frequency-BPE point in TPC. Positive $\Delta$BPC indicates Significance-Gain has lower BPC (better). Significance-Gain improves BPC at 4/5 matched points, with one degradation at $V{=}800$.}
\label{tab:matched_compression}
\end{table}

\paragraph{Discussion of regimes.}
Across most vocabulary sizes, Significance-Gain BPE achieves lower BPC at similar compression, suggesting that selecting merges by statistically cohesive association can improve downstream modeling beyond frequency-only compression. The degradation at $V{=}800$ indicates that the benefit is not uniform across operating points, motivating further analysis of score hyperparameters (e.g., $\alpha$) and merge inventory characteristics in future work.

\section{Limitations and Future Work}
\label{sec:limitations}

\paragraph{Tokenizer-dependent metrics.}
Per-token perplexity is not directly comparable across tokenizers; BPC (or bits-per-byte) should be treated as the primary metric for cross-tokenizer claims.

\paragraph{Single-run reporting.}
Results are from single runs per configuration (no multi-seed uncertainty). A complete evaluation should report mean$\pm$std over multiple seeds and include confidence intervals for BPC.

\paragraph{Objective and null model.}
The independence baseline captures adjacency cohesion but ignores longer-range structure. Understanding which merge types are favored (whitespace/punctuation vs.\ morphemes vs.\ domain strings) is an important direction for qualitative analysis.

\section{Conclusion}
\label{sec:conclusion}
Significance-Gain BPE replaces frequency-only merge selection with a statistically grounded cohesion term under an independence null, combined with an explicit compression gain factor. In a character-base WikiText-103 evaluation, the proposed tokenizer improves tokenizer-invariant BPC by about 0.9--1.0\% at a representative operating point and achieves mostly consistent gains under matched compression across a vocabulary-size sweep.

\section*{Code Availability}
The code to reproduce the tokenizer training, encoding pipeline, and language-model experiments is available at:
\url{https://github.com/Meetra21/LLM_24/blob/main/Significance_Gain_Pair_Encoding_for_LLMs_.ipynb}
\cite{nouri2025ijca}
\cite{nouri2025sobel}


\end{document}